\documentclass[sigconf,nonacm]{acmart}

\setcopyright{none}
\renewcommand\footnotetextcopyrightpermission[1]{}

\usepackage{amsmath}
\usepackage{booktabs}
\usepackage{graphicx}
\usepackage{microtype}
\usepackage{multirow}
\usepackage{pgfplots}
\usepackage{tikz}
\usetikzlibrary{arrows.meta,positioning}
\pgfplotsset{compat=1.18}
\AtBeginDocument{}

\title[VespaSeg]{VespaSeg: A Resource-Aware Ground-then-Segment Pipeline for Referring Expression Segmentation}

\author{Savindu Dilshan Wickramasinghe}
\affiliation{%
  \institution{Department of Electronic and Telecommunication Engineering, University of Moratuwa}
  \city{Moratuwa}
  \country{Sri Lanka}}
\email{savindudilshan124@gmail.com}

\begin{document}

\begin{abstract}
Referring expression segmentation requires language conditioned localization and pixel-accurate masks, but monolithic models can be costly to deploy. We present VespaSeg, a modular pipeline that grounds a text query with a compact vision--language model and converts the predicted box to a mask with MobileSAM. We study Florence-2-base, Florence-2-large, and Moondream2 grounders together with targeted adaptation of the grounding and segmentation stages. Under a repository-specific RefCOCO validation protocol containing the first expression for each of 3,811 referenced-object records, the adapted Florence-2-base pipeline obtains 73.64 mean intersection over union (mIoU) and 84.60 precision at IoU 0.5. On an NVIDIA RTX 6000 Ada GPU it processes 22.8 cached-image queries per second with 2.20~GB mean allocated GPU memory. A matched 500-query comparison gives 73.73 mIoU for Florence-2-base and 72.82 for Florence-2-large, while the base model is 1.70$\times$ faster and uses 1.17~GB less allocated memory. Ablations show that ground-truth-box adaptation raises MobileSAM mIoU from 82.22 to 86.61 and that reducing Florence-2's output-token budget from 64 to 32 preserves accuracy. These results support compact, modular grounding and segmentation, while also exposing the need for evaluation on the complete standard RefCOCO expression splits and deployment hardware.
\end{abstract}


\maketitle

\section{Introduction}
\label{sec:introduction}

Referring expression segmentation (RES) maps a natural-language expression and an image to the pixel mask of the described object. The task is useful wherever an interface must connect language to physical or visual entities, including robot manipulation and assistive interaction. Modern end-to-end RES systems have advanced rapidly, but their scale can complicate latency- and memory-constrained deployment.

VespaSeg explores a simple alternative: separate semantic localization from mask refinement. A vision--language model (VLM) first predicts a bounding box for an expression; MobileSAM~\cite{zhang2023mobilesam} then produces a mask from that box. The boundary between these stages makes it possible to exchange grounders, adapt each component independently, and cache the segmentation image embedding when multiple expressions refer to one image. We evaluate Florence-2~\cite{xiao2024florence} and Moondream2~\cite{moondream2025} as compact grounders and use low-rank adaptation (LoRA)~\cite{hu2022lora} where supported.

The experiments are reconstructed from the released training and evaluation artifacts. This audit changes the interpretation of two quantities from an earlier draft. First, the evaluation contains 3,811 instance--query pairs, formed by selecting the first expression from every RefCOCO validation record; it is therefore not directly comparable to results on all 10,834 validation expressions. Second, the reported rate measures sequential queries with MobileSAM image embeddings cached across repeated images, rather than independent end-to-end video frames. We retain these measurements because they characterize an interactive multi-query use case, but name and delimit them precisely.

The paper makes three contributions:

\begin{itemize}
  \item a modular ground-then-segment design using compact VLMs and MobileSAM;
  \item an artifact-backed comparison of accuracy, cached-query throughput, and allocated GPU memory, including a matched base-versus-large Florence-2 study; and
  \item component and inference-budget ablations, together with an explicit account of protocol limitations needed to reproduce and interpret the results.
\end{itemize}

\section{Related Work}
\label{sec:related}

\paragraph{Referring expression segmentation.}
RefCOCO introduced expression grounding in complex COCO scenes~\cite{yu2016refcoco,lin2014coco}. Early transformer-based RES approaches fused linguistic and visual representations directly. LAVT performs language-aware visual encoding~\cite{yang2022lavt}; CRIS transfers CLIP representations through text-to-pixel contrastive learning~\cite{wang2022cris}; and ReSTR formulates the task with convolution-free vision and language transformers~\cite{kim2022restr}. More recent systems join large language or multimodal models with segmentation decoders. LISA introduces a segmentation token whose embedding prompts SAM for reasoning segmentation~\cite{lai2024lisa}, while EVF-SAM studies early vision--language fusion for text-prompted SAM~\cite{zhang2024evfsam}. VespaSeg instead retains an explicit box interface between grounding and segmentation, prioritizing component interchangeability and measurable resource use.

\paragraph{Promptable and efficient segmentation.}
The Segment Anything Model (SAM) established promptable mask prediction at scale~\cite{kirillov2023sam}. MobileSAM replaces its heavy image encoder with TinyViT while retaining the prompt encoder and mask decoder~\cite{zhang2023mobilesam,wu2022tinyvit}. This makes box-prompted segmentation a practical second stage: the image embedding can be computed once, whereas new boxes require only the lighter prompt and mask decoders.

\paragraph{Compact visual grounding and adaptation.}
Florence-2 casts multiple vision tasks, including phrase grounding, into a prompt-based sequence-to-sequence interface and provides 231M- and 771M-parameter variants~\cite{xiao2024florence}. Moondream2 is a compact open model with a coordinate-returning pointing interface~\cite{moondream2025}. LoRA represents weight updates with trainable low-rank factors while freezing the original layer~\cite{hu2022lora}; this is useful when full VLM updates are too costly. VespaSeg compares LoRA-adapted Florence-2 grounders with a fully fine-tuned Moondream2 checkpoint and adapts selected MobileSAM mask-decoder projections with LoRA.

\section{Method}
\label{sec:method}

\subsection{Ground-then-Segment Inference}

\begin{figure}[t]
  \centering
  \resizebox{\linewidth}{!}{%
  \begin{tikzpicture}[
      node distance=3.4mm,
      every node/.style={font=\scriptsize,align=center},
      block/.style={draw,rounded corners,minimum height=7mm,minimum width=15mm,fill=gray!8},
      arrow/.style={-{Latex[length=1.7mm]},thick}]
    \node[block] (image) {image $I$};
    \node[block,below=of image] (text) {expression $q$};
    \node[block,right=of image,yshift=-5.2mm] (vlm) {compact VLM\\grounder};
    \node[block,right=of vlm] (box) {box $\hat b$};
    \node[block,right=of box] (sam) {MobileSAM};
    \node[block,right=of sam] (mask) {mask $\hat M$};
    \draw[arrow] (image.east) -- (vlm.west);
    \draw[arrow] (text.east) -- (vlm.west);
    \draw[arrow] (vlm) -- (box);
    \draw[arrow] (box) -- (sam);
    \draw[arrow] (sam) -- (mask);
    \draw[arrow] (image.north east) to[out=25,in=150] node[above,font=\tiny]{cached image embedding} (sam.north west);
  \end{tikzpicture}}
  \caption{VespaSeg separates language-conditioned box grounding from box-prompted mask prediction. MobileSAM's image embedding can be reused for multiple expressions on the same image.}
  \Description{A flow diagram. An image and a text expression enter a compact vision-language grounder, which outputs a box. The box and a cached image embedding enter MobileSAM, which outputs a mask.}
  \label{fig:pipeline}
\end{figure}
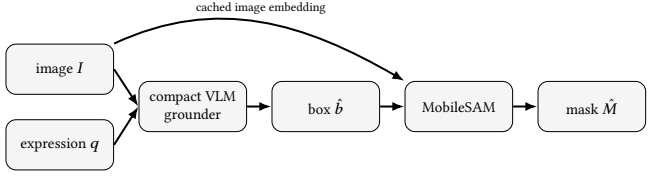

For image $I$ and expression $q$, a grounder $G_\theta$ predicts a box
\begin{equation}
  \hat{b}=G_\theta(I,q)=(x_1,y_1,x_2,y_2).
\end{equation}
MobileSAM first encodes the image as $z=E_\phi(I)$. Its prompt encoder and mask decoder then produce
\begin{equation}
  \hat{M}=D_\psi\!\left(z,P(\hat{b})\right).
\end{equation}
The image embedding $z$ is reused for consecutive queries on the same image. Figure~\ref{fig:pipeline} summarizes this interface.

Florence-2 receives the \texttt{<CAPTION\_TO\_PHRASE\_GROUNDING>} task token followed by $q$ and returns coordinates that are decoded to the input image frame. Moondream2 receives a pointing query and returns normalized coordinates, which are likewise converted to a box. Invalid or unordered coordinates are clamped before being passed to MobileSAM.

\subsection{Adaptation}

For a linear projection $W\in\mathbb{R}^{d_o\times d_i}$, LoRA uses
\begin{equation}
  W' = W + \frac{\alpha}{r}BA,
\end{equation}
where $A\in\mathbb{R}^{r\times d_i}$ and $B\in\mathbb{R}^{d_o\times r}$ are trainable. Florence-2 LoRA targets the \texttt{q\_proj}, \texttt{k\_proj}, \texttt{v\_proj}, \texttt{out\_proj}, \texttt{fc1}, and \texttt{fc2} projections. The evaluated Florence-2-base checkpoint uses $r=8$, $\alpha=16$, and the epoch-2 state from a three-epoch run. The Florence-2-large checkpoint uses the same rank and scale after three epochs. Training uses every available expression, yielding 120,624 training pairs, a batch size of 4, gradient accumulation of 4, and AdamW~\cite{loshchilov2019adamw} with a learning rate of $10^{-4}$.

MobileSAM LoRA is applied to eligible mask-decoder linear layers with $r=8$ and $\alpha=16$. It trains on 42,404 RefCOCO training records for 10 epochs with batch size 4 and learning rate $10^{-4}$. Ground-truth boxes isolate mask adaptation from grounding error. The loss is
\begin{equation}
  \mathcal{L}_{\mathrm{mask}}=\mathcal{L}_{\mathrm{Dice}}+0.5\,\mathcal{L}_{\mathrm{BCE}}.
\end{equation}
This updates 167,936 of 10,298,028 parameters (1.63\%). The Moondream2 checkpoint was fully fine-tuned, rather than LoRA-adapted, on the first expression of each of the 42,404 training records for three epochs with batch size 2 and learning rate $10^{-5}$.

\section{Experimental Setup}
\label{sec:setup}

\subsection{Data and Evaluation Protocol}

RefCOCO contains COCO images, referred objects, and multiple expressions per object~\cite{yu2016refcoco,lin2014coco}. Training and validation use the Hugging Face representation \texttt{jxu124/refcoco}. Florence-2 training expands every sentence in the 42,404 training records to 120,624 expression pairs. The MobileSAM and Moondream2 training scripts use one expression per training record.

The released evaluation loop selects sentence index zero for each validation record, producing 3,811 instance--query pairs. The standard RefCOCO validation split contains 10,834 expressions for these records. Consequently, the reported values describe the repository protocol and must not be placed in a standard RefCOCO leaderboard without a complete re-evaluation. No testA or testB results are claimed.

For a predicted mask $\hat M_i$ and ground truth $M_i$, we report
\begin{equation}
  \operatorname{mIoU}=\frac{1}{N}\sum_{i=1}^{N}
  \frac{|\hat M_i\cap M_i|}{|\hat M_i\cup M_i|}
\end{equation}
and $\mathrm{P@0.5}=N^{-1}\sum_i\mathbb{1}[\operatorname{IoU}_i\geq0.5]$. Both are presented as percentages.

\subsection{Runtime Measurement}

Measurements were collected on one NVIDIA RTX 6000 Ada Generation GPU using half-precision inference and PyTorch scaled dot-product attention. Unless stated otherwise, Florence-2 receives images resized to a maximum side of 320 pixels and may generate at most 32 tokens. Five samples warm up the pipeline.

The timed interval includes VLM preprocessing and generation plus box-prompted MobileSAM decoding. MobileSAM's image encoder is run outside the per-query timer when the input image changes, and its embedding is reused for consecutive queries to that image. We therefore report \emph{cached-image query throughput} in queries per second (q/s), not video frame rate. Memory is the mean PyTorch allocated GPU memory sampled after each query; it is not total board consumption. These definitions apply to the Florence results for which compatible profiling artifacts were retained. Moondream2 accuracy is reported without a cross-run resource comparison because the available logs used different memory and instrumentation paths.

\section{Results}
\label{sec:results}

\subsection{Repository-Protocol Accuracy}

Table~\ref{tab:full} reports the runs that cover all 3,811 pairs in the repository protocol. Adapted Florence-2-base gives the strongest result, reaching 73.64 mIoU and 84.60 P@0.5. The unadapted Florence rows already use the adapted MobileSAM; their lower scores therefore primarily expose grounding error. The two Moondream2 rows change both the grounder and MobileSAM and should not be read as a controlled single-component ablation.

\begin{table*}[t]
  \caption{Results on the 3,811-pair repository-specific RefCOCO validation protocol. ``Adaptation'' names the stages changed from their released base weights. Query throughput and allocated memory are reported only for compatible Florence profiling runs. Best accuracy is bold.}
  \label{tab:full}
  \centering
  \small
  \begin{tabular}{llrrrrr}
    \toprule
    Grounder & Adaptation & Pairs & mIoU (\%) & P@0.5 (\%) & q/s $\uparrow$ & Alloc. GB $\downarrow$ \\
    \midrule
    Florence-2-base  & MobileSAM LoRA                    & 3,811 & 40.40 & 43.32 & 15.3 & 2.27 \\
    Florence-2-large & MobileSAM LoRA                    & 3,811 & 43.28 & 47.39 &  9.7 & 3.38 \\
    Moondream2       & none                              & 3,811 & 62.23 & 69.56 & --   & --   \\
    Moondream2       & grounder full FT + MobileSAM LoRA & 3,811 & 63.22 & 71.37 & --   & --   \\
    Florence-2-base  & grounder LoRA + MobileSAM LoRA    & 3,811 & \textbf{73.64} & \textbf{84.60} & 22.8 & 2.20 \\
    \bottomrule
  \end{tabular}
\end{table*}

The 33.24-point difference between the adapted and unadapted Florence-2-base systems is large, but it is not an unbiased training-effect estimate: the retained checkpoint was selected during development on validation behavior, and no multiple-seed or held-out test result is available. The comparison nevertheless indicates that RefCOCO-specific grounding is the principal source of improvement in the evaluated pipeline.

\subsection{Matched Florence-2 Resource Comparison}

Table~\ref{tab:matched} compares adapted Florence variants on the same first 500 validation pairs. The base model is slightly more accurate in this subset, processes 1.70$\times$ as many cached-image queries per second, and uses 1.17~GB less allocated memory. Thus, increasing grounder capacity does not improve this particular operating point. This is a within-system comparison, not evidence that the smaller model universally outperforms larger grounders.

\begin{table}[t]
  \caption{Matched 500-pair comparison of adapted Florence-2 variants with adapted MobileSAM.}
  \label{tab:matched}
  \centering
  \small
  \begin{tabular}{lrrrr}
    \toprule
    Grounder & mIoU & P@0.5 & q/s & GB \\
    \midrule
    Florence-2-base  & \textbf{73.73} & \textbf{85.00} & \textbf{22.2} & \textbf{2.21} \\
    Florence-2-large & 72.82 & 83.40 & 13.0 & 3.38 \\
    \bottomrule
  \end{tabular}
\end{table}

\subsection{Component and Inference-Budget Ablations}

Using ground-truth boxes removes grounding errors and isolates MobileSAM. On a fixed 1,000-record validation subset, mask-decoder LoRA improves mIoU from 82.22 to 86.61 (Table~\ref{tab:components}). Separately, full Moondream2 fine-tuning improves box IoU from 69.33 to 70.09 on a 500-record validation subset. These tasks and subset sizes differ, so only the within-row changes are meaningful.

\begin{table}[t]
  \caption{Component-level adaptation with ground-truth targets. $N$ is the number of validation records.}
  \label{tab:components}
  \centering
  \small
  \begin{tabular}{lrrr}
    \toprule
    Component metric & $N$ & Before & After \\
    \midrule
    MobileSAM mask mIoU (\%) & 1,000 & 82.22 & \textbf{86.61} \\
    Moondream2 box IoU (\%)  &   500 & 69.33 & \textbf{70.09} \\
    \bottomrule
  \end{tabular}
\end{table}

Table~\ref{tab:budget} varies the Florence-2-base input resolution and generation budget across all 3,811 pairs. Cutting the token limit from 64 to 32 changes neither accuracy nor memory and slightly increases throughput. Reducing the maximum image side to 256 or 192 pixels brings only small throughput gains and progressively lowers mIoU. We therefore use 320 pixels and 32 tokens as the reported operating point.

\begin{table}[t]
  \caption{Input and generation-budget sensitivity for adapted Florence-2-base + adapted MobileSAM ($N=3{,}811$).}
  \label{tab:budget}
  \centering
  \small
  \begin{tabular}{rrrrr}
    \toprule
    Max side & Tokens & mIoU & P@0.5 & q/s \\
    \midrule
    320 & 64 & 73.64 & 84.60 & 22.50 \\
    320 & 32 & \textbf{73.64} & \textbf{84.60} & 22.78 \\
    256 & 64 & 73.54 & 84.41 & 22.85 \\
    256 & 32 & 73.54 & 84.41 & 22.91 \\
    192 & 32 & 72.76 & 83.55 & \textbf{23.10} \\
    \bottomrule
  \end{tabular}
\end{table}

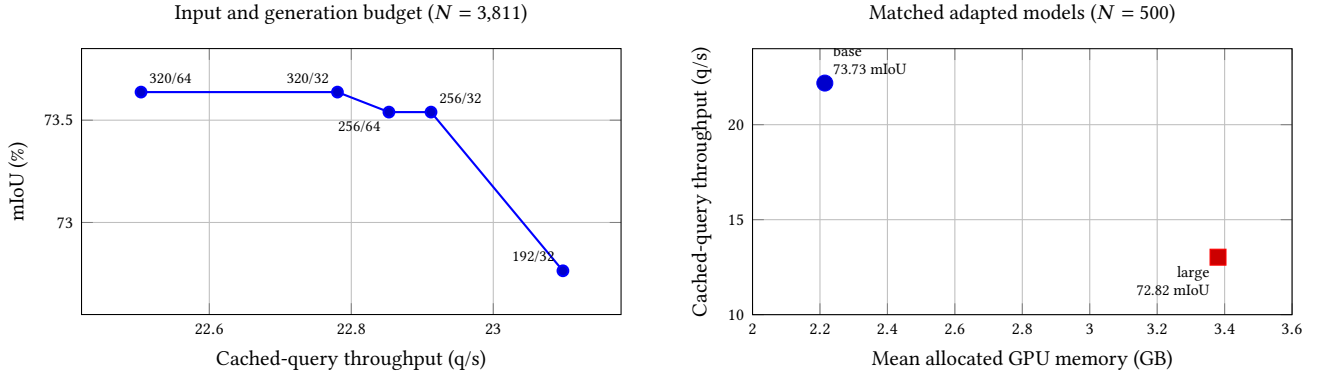
\begin{figure*}[t]
  \centering
  \begin{minipage}[t]{0.49\textwidth}
    \centering
    \begin{tikzpicture}
      \begin{axis}[
          width=\linewidth,
          height=5.1cm,
          xmin=22.42,xmax=23.18,
          ymin=72.55,ymax=73.85,
          xlabel={Cached-query throughput (q/s)},
          ylabel={mIoU (\%)},
          grid=both,
          tick label style={font=\scriptsize},
          label style={font=\small},
          title={Input and generation budget ($N=3{,}811$)},
          title style={font=\small},
          legend style={font=\scriptsize,draw=none,at={(0.02,0.02)},anchor=south west}]
        \addplot+[mark=*,thick,blue]
          coordinates {(22.5038,73.6373) (22.7805,73.6373) (22.8530,73.5395) (22.9123,73.5395) (23.0981,72.7638)};
        \node[font=\scriptsize,anchor=south west] at (axis cs:22.5038,73.6373) {320/64};
        \node[font=\scriptsize,anchor=south east] at (axis cs:22.7805,73.6373) {320/32};
        \node[font=\scriptsize,anchor=north east] at (axis cs:22.8530,73.5395) {256/64};
        \node[font=\scriptsize,anchor=south west] at (axis cs:22.9123,73.5395) {256/32};
        \node[font=\scriptsize,anchor=south east] at (axis cs:23.0981,72.7638) {192/32};
      \end{axis}
    \end{tikzpicture}
  \end{minipage}\hfill
  \begin{minipage}[t]{0.49\textwidth}
    \centering
    \begin{tikzpicture}
      \begin{axis}[
          width=\linewidth,
          height=5.1cm,
          xmin=2.0,xmax=3.6,
          ymin=10,ymax=24,
          xlabel={Mean allocated GPU memory (GB)},
          ylabel={Cached-query throughput (q/s)},
          grid=both,
          tick label style={font=\scriptsize},
          label style={font=\small},
          title={Matched adapted models ($N=500$)},
          title style={font=\small}]
        \addplot+[only marks,mark=*,mark size=3pt,blue] coordinates {(2.2144,22.1899)};
        \addplot+[only marks,mark=square*,mark size=3pt,red] coordinates {(3.3804,13.0255)};
        \node[font=\scriptsize,anchor=south west,align=left] at (axis cs:2.2144,22.1899) {base\\73.73 mIoU};
        \node[font=\scriptsize,anchor=north east,align=right] at (axis cs:3.3804,13.0255) {large\\72.82 mIoU};
      \end{axis}
    \end{tikzpicture}
  \end{minipage}
  \caption{Artifact-backed resource trade-offs. Left: the five non-compiled configurations in \texttt{florence\_base\_optimization.json}; point labels are maximum image side/output-token limit. Right: the adapted Florence-2 variants in \texttt{florence\_4gb\_benchmark.json}. The plotted rates use the cached-image timing protocol in Section~\ref{sec:setup}.}
  \Description{Two plots. The left plot shows mean IoU decreasing slightly as cached-query throughput increases through smaller image sizes. The right plot shows Florence-2-base at lower allocated memory and higher throughput than Florence-2-large, with similar mean IoU.}
  \label{fig:tradeoffs}
\end{figure*}

Figure~\ref{fig:tradeoffs} visualizes the two resource comparisons tabulated above. On the full protocol, the frontier is shallow: reducing the generation budget is effectively free, whereas the 192-pixel input trades 0.88 mIoU points for only 0.32 additional q/s relative to the selected 320/32 setting. In the matched model-size comparison, Florence-2-base lies above and to the left of Florence-2-large, indicating higher cached-query throughput with lower allocated memory on that subset.

\subsection{Qualitative Observations}

\begin{figure*}[t]
  \centering
  \includegraphics[width=0.86\textwidth]{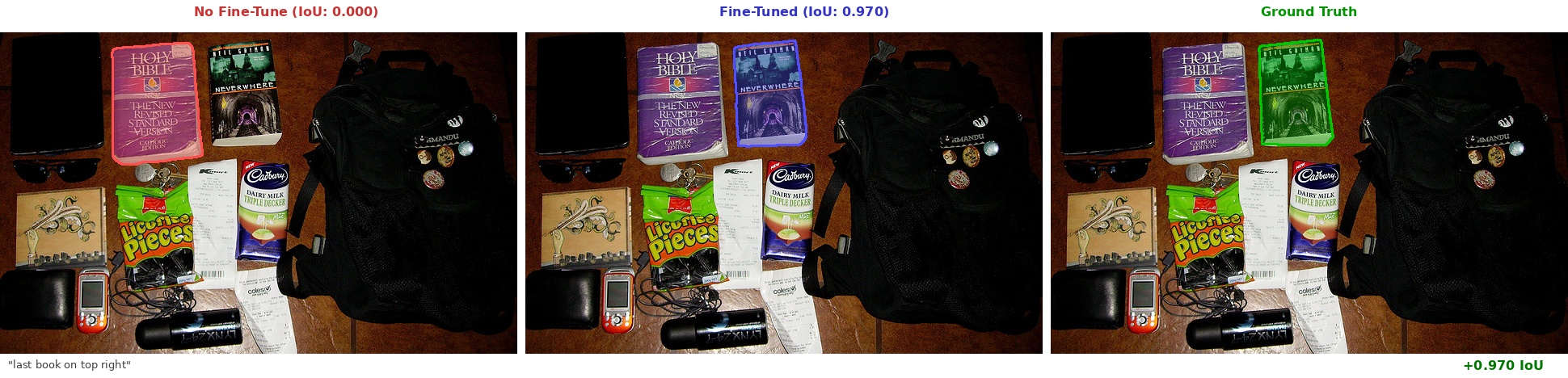}\\[1.2mm]
  \includegraphics[width=0.86\textwidth]{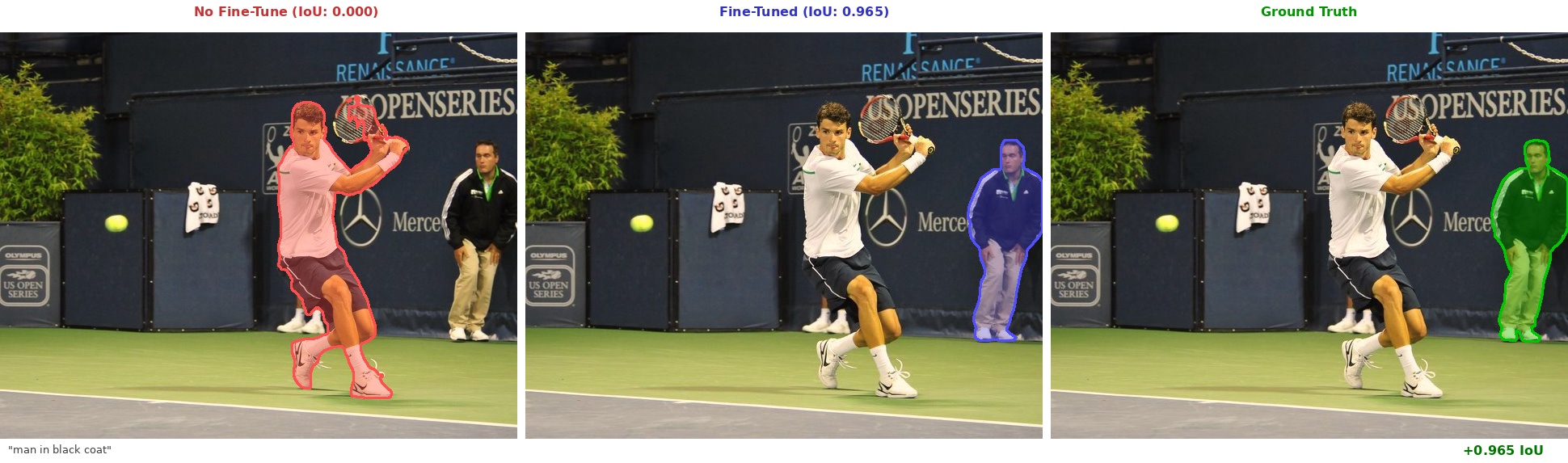}\\[1.2mm]
  \includegraphics[width=0.86\textwidth]{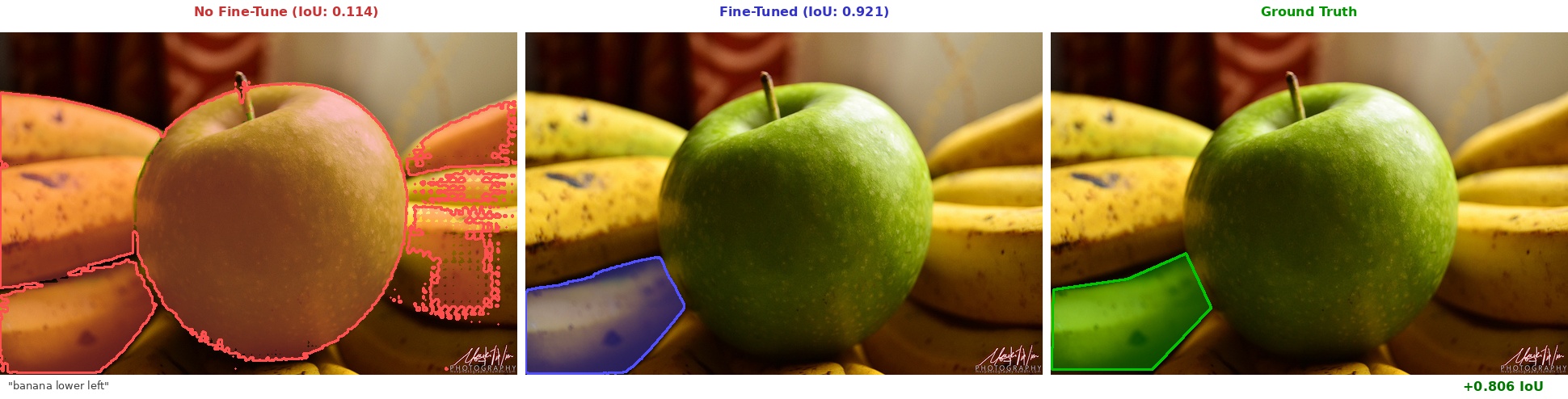}
  \caption{Selected high-improvement RefCOCO validation examples from a 200-pair diagnostic subset. Each montage shows the unadapted prediction (left), adapted prediction (middle), and ground truth (right); the expression and exact IoUs are embedded in the artifact. From top to bottom, IoU changes are $0.000\!\to\!0.970$, $0.000\!\to\!0.965$, and $0.114\!\to\!0.921$.}
  \Description{Three comparison rows. Adaptation changes the selected object to the requested top-right book, man in a black coat, and lower-left banana. Each row contrasts unadapted, adapted, and ground-truth masks.}
  \label{fig:qualitative}
\end{figure*}

Figure~\ref{fig:qualitative} adds three stored comparisons from a 200-pair diagnostic subset. Two complete grounding failures recover from zero IoU to 0.970 and 0.965: the adapted model follows a spatial book description and a clothing-based person description, respectively. A partial error improves from 0.114 to 0.921 when the model changes its selection from the apple region to the requested lower-left banana. These cases were selected for large improvement, are not representative samples, and do not replace aggregate evaluation.

\section{Limitations and Reproducibility}
\label{sec:limitations}

The current evidence is preliminary. Most importantly, evaluation uses only the first expression in each validation record rather than every standard RefCOCO expression, and it omits testA and testB. Development and checkpoint selection also used validation behavior, so the reported accuracy may be optimistic. Multiple seeds, uncertainty intervals, and significance tests were not recorded. Some component comparisons change more than one stage, and the large Florence adapted run has only a 500-pair resource evaluation.

The throughput experiment benefits from cached MobileSAM image embeddings and was conducted on a high-end desktop GPU. It therefore does not establish video-frame throughput, embedded-device latency, energy use, or deployability within a 2.20~GB physical-memory budget. PyTorch allocated memory excludes parts of the runtime and driver. Finally, grounding errors are unrecoverable by the box-prompted mask decoder, while a modular pipeline does not jointly optimize language reasoning and boundary prediction. Future work should re-run the complete standard splits on held-out checkpoints, add independent end-to-end latency and peak-memory measurements on edge hardware, and compare against current RES systems under the same protocol.

The code, training scripts, and retained result artifacts are available at \url{https://github.com/Savidilsh/VespaSeg}. The repository should be treated as the source of the custom protocol rather than as a drop-in implementation of the standard RefCOCO leaderboard protocol.

\section{Conclusion}
\label{sec:conclusion}

VespaSeg demonstrates that an explicit interface between visual grounding and promptable segmentation can balance accuracy and resource use. On the audited 3,811-pair protocol, adapted Florence-2-base with MobileSAM reaches 73.64 mIoU and 84.60 P@0.5, and a matched subset shows a clear resource advantage over Florence-2-large. Component ablations attribute gains to both grounding and mask adaptation, while generation-budget tests identify a no-loss reduction from 64 to 32 output tokens. These findings motivate compact modular RES, but standard-split and edge-device evaluation remain necessary before making broader state-of-the-art or real-time deployment claims.

\bibliographystyle{ACM-Reference-Format}
\bibliography{references}

\end{document}